\documentclass[11pt,a4paper]{article}

\usepackage[utf8]{inputenc}
\usepackage[T1]{fontenc}
\usepackage{lmodern}
\usepackage{authblk}
\usepackage{microtype}
\usepackage{graphicx}
\usepackage{xcolor}
\usepackage{booktabs}
\usepackage{hyperref}
\usepackage[margin=1in]{geometry}
\usepackage{titlesec}
\usepackage{enumitem}
\usepackage{fancyhdr}
\usepackage{float}
\usepackage{amsmath}
\usepackage{tikz}

\usetikzlibrary{arrows.meta}

\usepackage{pgfplots}
\usepackage{algorithm}
\usepackage{algpseudocode}
\usepackage[style=authoryear,natbib=true]{biblatex}
\definecolor{primary}{RGB}{0, 83, 155}
\definecolor{secondary}{RGB}{105, 165, 203}
\definecolor{accent}{RGB}{255, 102, 0}

\hypersetup{
    colorlinks=true,
    linkcolor=primary,
    filecolor=primary,
    urlcolor=primary,
    citecolor=primary
}

\titleformat{\section}
  {\normalfont\Large\bfseries\color{primary}}
  {\thesection}{1em}{}
\titleformat{\subsection}
  {\normalfont\large\bfseries\color{primary}}
  {\thesubsection}{1em}{}

\title{\Huge\textbf{\textcolor{primary}{Transparent AI: The Case for Interpretability and Explainability}}}

\author[1]{Dhanesh Ramachandram}
\author[1]{Himanshu Joshi}
\author[1]{Judy Zhu}
\author[1]{Dhari Gandhi}
\author[1,3]{Lucas Hartman}
\author[1]{Ananya Raval}

\affil[1]{Vector Institute for Artificial Intelligence, Toronto}
\affil[2]{University of Western Ontario}
\date{\today}

\begin{document}

\maketitle
\thispagestyle{empty}

\begin{abstract}
As artificial intelligence systems increasingly inform high-stakes decisions across sectors, transparency has become foundational to responsible and trustworthy AI implementation. Leveraging our role as a leading institute in advancing AI research and enabling industry adoption, we present key insights and lessons learned from practical interpretability applications across diverse domains. This paper offers actionable strategies and implementation guidance tailored to organizations at varying stages of AI maturity, emphasizing the integration of interpretability as a core design principle rather than a retrospective add-on.
\end{abstract}

\newpage
\tableofcontents
\newpage


\section{Introduction}
\subsection{Background and Motivation}
The increasing reliance on artificial intelligence (AI) and machine learning (ML) systems in critical sectors such as healthcare, finance, and public administration highlights the importance of transparency in their decision-making processes \cite{doshi2017considerations}. A central challenge faced in deploying these technologies involves balancing model performance and interpretability. While advanced models such as deep neural networks provide exceptional predictive capabilities, their complexity often results in a lack of transparency, commonly referred to as “black box” behavior \cite{nauta2023systematic}.

Consider, for instance, a medical diagnostic tool that boasts a mere 2\% error rate but offers no explanation of its predictions compared to a human physician whose decisions have a 15\% error rate but can be fully explained. Most stakeholders would hesitate to trust a system that cannot justify its decisions despite superior performance metrics. This example underscores the critical balance between performance and interpretability, a core concern in the ethical and practical deployment of AI.

The demand for explainable and interpretable AI goes beyond academic interest and is being increasingly recognized as essential for the responsible and ethical use of AI in sectors directly impacting human health, safety, and legal rights. As highlighted by \citet{doshi2017considerations}, interpretability becomes especially crucial in situations where incorrect predictions have significant consequences, stakeholder justification is required, or regulatory frameworks demand transparency. Examples of such domains are as follows: 

\paragraph{Healthcare Applications:} In healthcare, the importance of interpretability cannot be overstated. Clinicians, such as radiologists employing AI tools for cancer detection, require clarity not only on the prediction itself but also on the specific factors influencing that prediction. Without this insight, healthcare professionals cannot confidently integrate AI recommendations with clinical judgment, potentially risking patient outcomes through misdiagnoses or incorrect treatments. Moreover, patients and their families possess a fundamental right to clear explanations about medical decisions affecting their care. The interpretability challenge intensifies in urgent medical scenarios. 

\paragraph{Financial Services:} In finance, interpretability addresses both regulatory compliance and ethical accountability. For example, financial institutions must justify credit decisions clearly and transparently to comply with fair lending regulations. According to \citet{nauta2023systematic}, while 58\% of studies on explainable AI involve quantitative evaluations, many do not adequately meet the stringent transparency requirements imposed by financial regulations, which demand accurate, reliable, and legally defensible explanations.

\paragraph{Public Sector and Justice:} Governmental use of algorithmic decision-making for allocating resources, determining eligibility for benefits, and criminal justice applications places interpretability at the forefront of public accountability. Citizens expect clear explanations for decisions impacting their lives, highlighting the democratic necessity of interpretable AI in the public sector. Transparency in these decisions not only supports fairness but also reinforces trust in public institutions.

\subsection{Scope and Objectives}
Given these challenges and imperatives, this white paper provides a high-level overview of interpretability and explainable AI, outlines essential principles for achieving transparency, discusses domain-specific considerations for effective explanations, reviews suitable evaluation methods, and offers practical insights to help organizations integrate interpretable AI responsibly into their decision-making processes. Lessons learned from conducting industry-focused interpretability application on this topic with diverse participation from the Canadian AI ecosystem are also presented.
It is hoped that this whitepaper will help drive adoption of explainability and interpretable models across various industries by providing some adoption guidelines.

\section{Fundamentals of Interpretability and Explainability}

\subsection{Definitions and Key Concepts}
\begin{figure}[ht]
    \centering
    \includegraphics[scale=0.3]{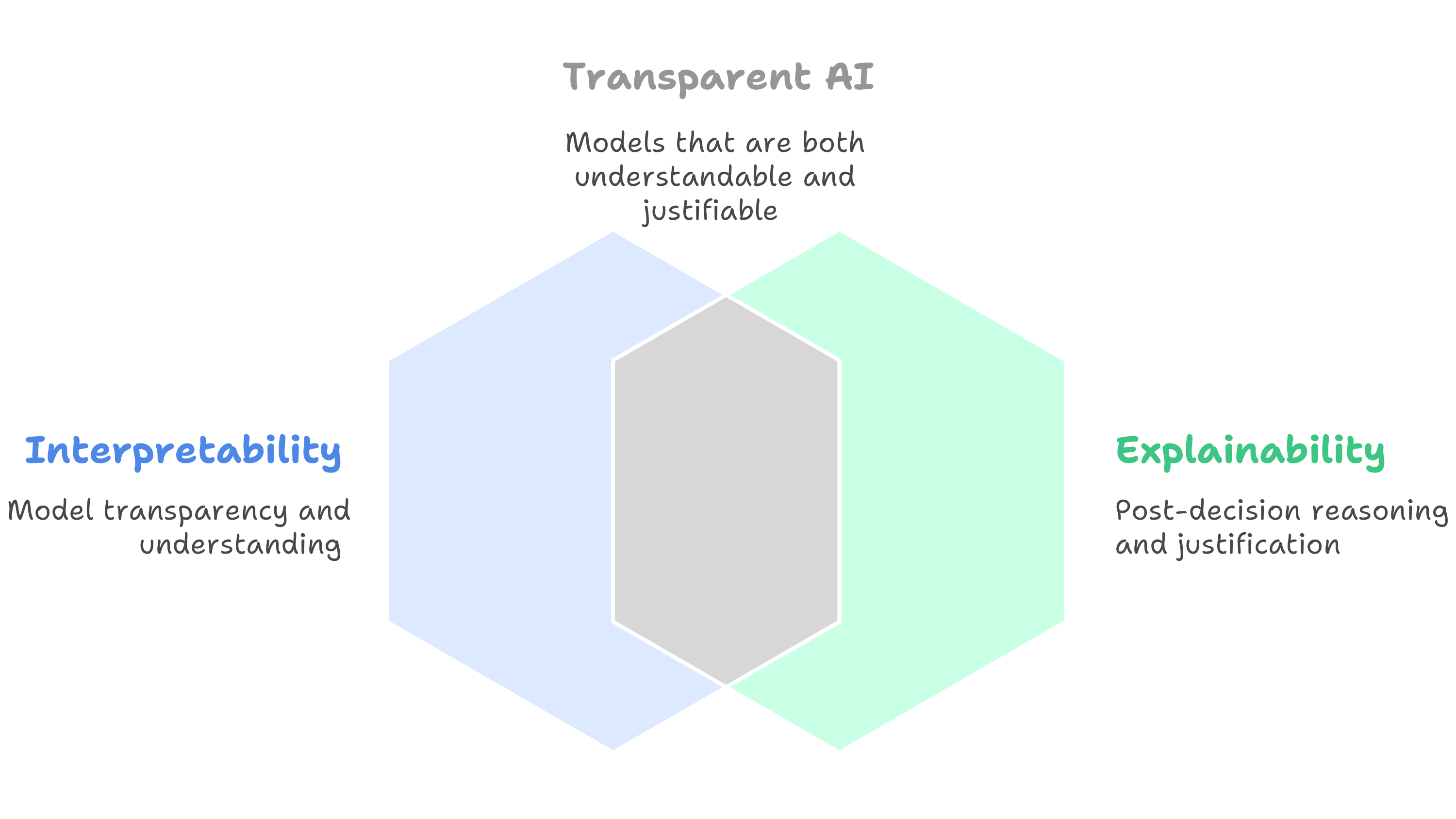}
    \caption{Relationship Between Interpretability, Explainability, and Transparent AI}
    \label{fig:transparency}
\end{figure}
Interpretability refers to how easily a human can understand the logic and decision-making processes within a machine learning model. Explainability, in contrast, pertains to the ability to provide clear and understandable reasons for specific decisions made by a model after those decisions have been made (Fig.~\ref{fig:transparency}).

In practical terms:
\begin{itemize}
\item \textbf{Interpretability} addresses the question: \textit{“How does the model function internally?”}
\item \textbf{Explainability} addresses the question: \textit{“Why did the model make a particular decision?”}
\end{itemize}

Interpretable or \textit{glass-box} models, such as decision trees and linear regression, are inherently transparent and understandable by design. In contrast, Explainable AI (XAI) typically involves supplementary techniques aimed at making the outputs and decisions of more complex, \textit{black-box} models, such as neural networks, understandable and justifiable to users.

\begin{figure}[ht]
  \centering
  \begin{tikzpicture}
    \begin{axis}[
      width=0.9\textwidth,
      height=0.6\textwidth,
      xmin=0.5, xmax=10.5,
      ymin=-0.5, ymax=10.5,
      xtick={1,9},
      xticklabels={Low,High},
      ytick={0,9},
      yticklabels={Low,High},
      xlabel={Model Interpretability},
      ylabel={Model Complexity \& Performance},
      title={Model Interpretability vs.\ Performance},
      grid=both,
      grid style={gray!30},
      axis lines=left,
      clip=false
    ]
    \addplot[smooth, thick, blue] coordinates {
  (1,9.00) (2,4.50) (3,3.00) (4,2.25)
  (5,1.80) (6,1.50) (7,1.29) (8,1.12)
  (9,1.00)
};

      \addplot[only marks, mark=*, mark options={fill=blue}] coordinates {
          (1,9.00) (2,4.50) (3,3.00) (4,2.25)
  (5,1.80) (6,1.50) (7,1.29) (8,1.12)
  (9,1.00)
      };

      \node[anchor=west] at (axis cs:1,9) {Deep Learning \& LLMs};
      \node[anchor=south] at (axis cs:2,4.5) {Ensemble Models};
      \node[anchor=south] at (axis cs:3,3) {MLPs};
      \node[anchor=west] at (axis cs:4,2.5) {SVMs};
      \node[anchor=east] at (axis cs:5,1.8) {Bayesian Models};
      \node[anchor=north] at (axis cs:6,1.5) {GAMs};
      \node[anchor=south] at (axis cs:7,1.29) {Logistic Regression};
      \node[anchor=north] at (axis cs:8,1.12) {Decision Trees};
      \node[anchor=west] at (axis cs:9,1) {Rule‑based Learning};

\addplot[only marks, mark=o, mark options={ultra thick,draw=black,fill=white}]
  coordinates {(8,8)};
\node[anchor=west] at (axis cs:8,8) {Ideal solution};

\draw[->, very thick,red]
  (axis cs:4,4) -- (axis cs:8,8)
  node[midway, sloped, above] {Interpretable ML research};
    \end{axis}
  \end{tikzpicture}
  \caption{Trade‑off Between Model Interpretability and Learning Performance.}
  \label{fig:complex}
\end{figure}
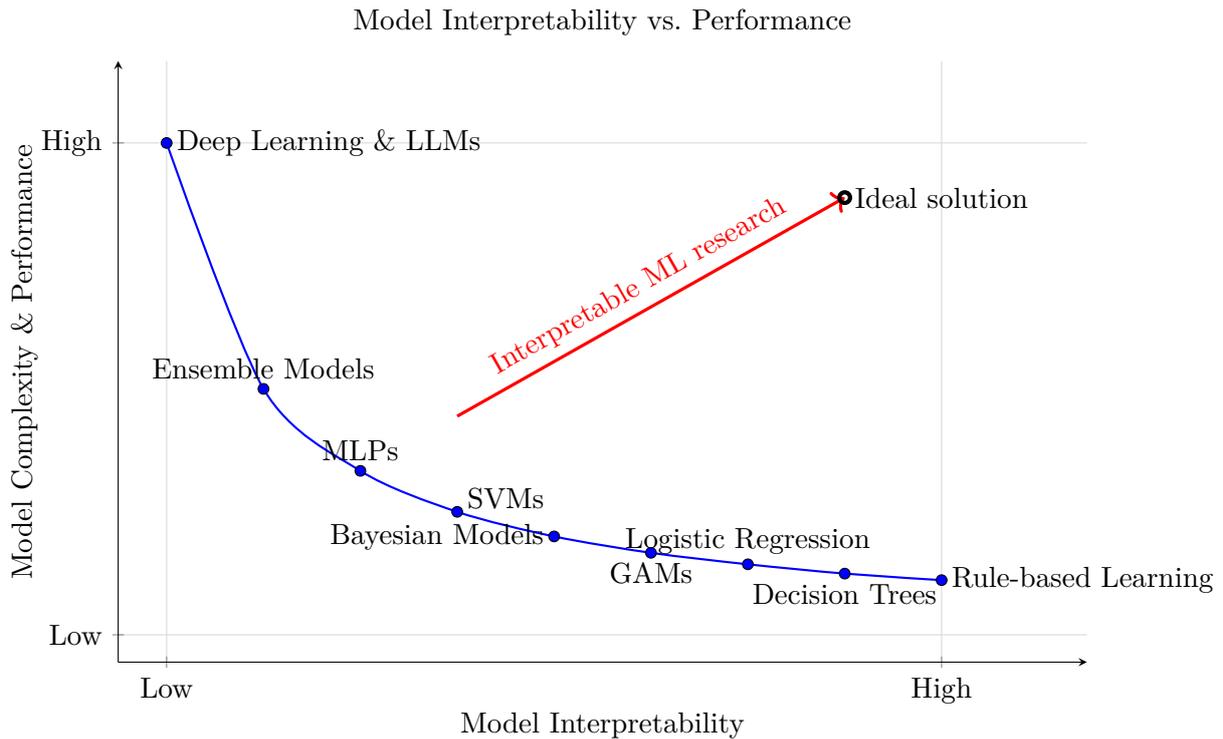

With the rapid advancements in artificial intelligence, particularly in deep learning and generative AI technologies, models have become increasingly complex. This growing complexity further underscores the critical need for transparency and interpretability in AI systems. In response to this growing need, a focused area of scientific research has emerged, aimed at developing high-performing AI models that are inherently interpretable. This direction, illustrated in Figure~\ref{fig:complex}, seeks to bridge the gap between model accuracy and transparency.

\subsection{Taxonomy of Methods}

Explanation and interpretability methods can be broadly categorized along several dimensions to help practitioners choose appropriate tools for their use case. A common distinction is between \textit{model-specific} and \textit{model-agnostic} methods. Model-specific techniques exploit the internal structure of a given model class and are generally more precise, whereas model-agnostic approaches treat the model as a black box and can be applied to a wider range of architectures.

Another useful categorization is \textit{local} versus \textit{global} explanations. Local explanations focus on individual predictions, helping users understand why a specific output was produced. Global explanations describe the overall behavior of the model, providing insights into general trends and feature influences.

Some methods also vary by whether they are \textit{post hoc} applied after model training or whether they are built into the model itself (\textit{inherently interpretable}). Post hoc methods can be useful for auditing and retrospective analysis, while inherently interpretable models are often designed with transparency from the outset. Understanding these distinctions is essential for selecting techniques that align with organizational goals, domain requirements, and model constraints. 
It is not the aim of this whitepaper to provide technical descriptions of explainability and interpretable models and for that we refer readers to recent publications~\citep{molnar2020interpretable,hsieh2024,luo2024,kamath2021,černevičienė2024,kenesei2015}
In the next section, we discuss the current policy and regulatory landscape in the context of explainability and interpretability

\section{Policy Landscape and Regulatory Considerations}

Governments worldwide are implementing policies that mandate AI interpretability, acknowledging that transparent algorithms are necessary to safeguard individual rights and enable accountable governance. Fo example, the European Union's proposed Artificial Intelligence Act explicitly requires transparency and explainability for high-risk AI applications, establishing legal obligations for AI system providers to ensure their systems are interpretable by users and affected parties. Similarly, emerging guidelines from healthcare regulators, including the Food and Drug Administration (FDA)'s guidance on AI/ML-based medical devices, emphasize the need for interpretable AI in medical applications where patient safety is paramount. These regulatory developments reflect a paradigm shift where interpretability is not merely a technical preference but a fundamental requirement for ethical AI deployment in society.

Multiple jurisdictions have established comprehensive AI interpretability requirements across general and sector-specific regulations.  Canada's proposed AIDA emphasizes risk-based governance with interpretability assessments during development phases. Sector-specific regulations include FDA guidance requiring clear documentation of AI/ML medical device decision-making and financial regulators expecting interpretable AI-driven risk models under frameworks like Basel III.
The EU AI Act mandates transparency for high-risk AI systems and grants individuals the right to clear and meaningful explanations of algorithmic decisions. The US White House Blueprint for AI Bill of Rights (2022) establishes interpretability as a fundamental civil right, requiring notice and explanation for impactful algorithmic systems.

These frameworks converge on several core demands: risk-based transparency requirements where higher-risk applications face stricter interpretability standards, mandatory human oversight capabilities, and individual rights to understand algorithmic decisions affecting them. Practitioners can fulfill these requirements by designing interpretability into systems from the development phase rather than retrofitting, providing clear documentation of decision-making processes, conducting impact assessments to determine appropriate interpretability levels for specific use cases, and ensuring explanations are accessible and understandable to affected parties. The key shift is treating interpretability as a fundamental design requirement rather than an optional technical feature.

Although many regulations call for transparency, few provide clear definitions of what constitutes``sufficient'' explanation. Standardizing definitions and evaluation metrics is an ongoing challenge. Most regulatory documents do not explicitly specify how explainability of models should be evaluated, nor what metrics should be used and how these results should be reported. This is possibly due to the multitude of AI-driven applications that exists and may require case-specific evaluation criteria.  

Next,  we discuss stakeholder specific guidelines pertaining to explainability of ML models and how these stakeholders can play unique roles in furthering this cause for responsible AI.
\section{Stakeholder-Specific Guidelines}
\begin{figure}[H]
    \centering
    \includegraphics[scale=0.3]{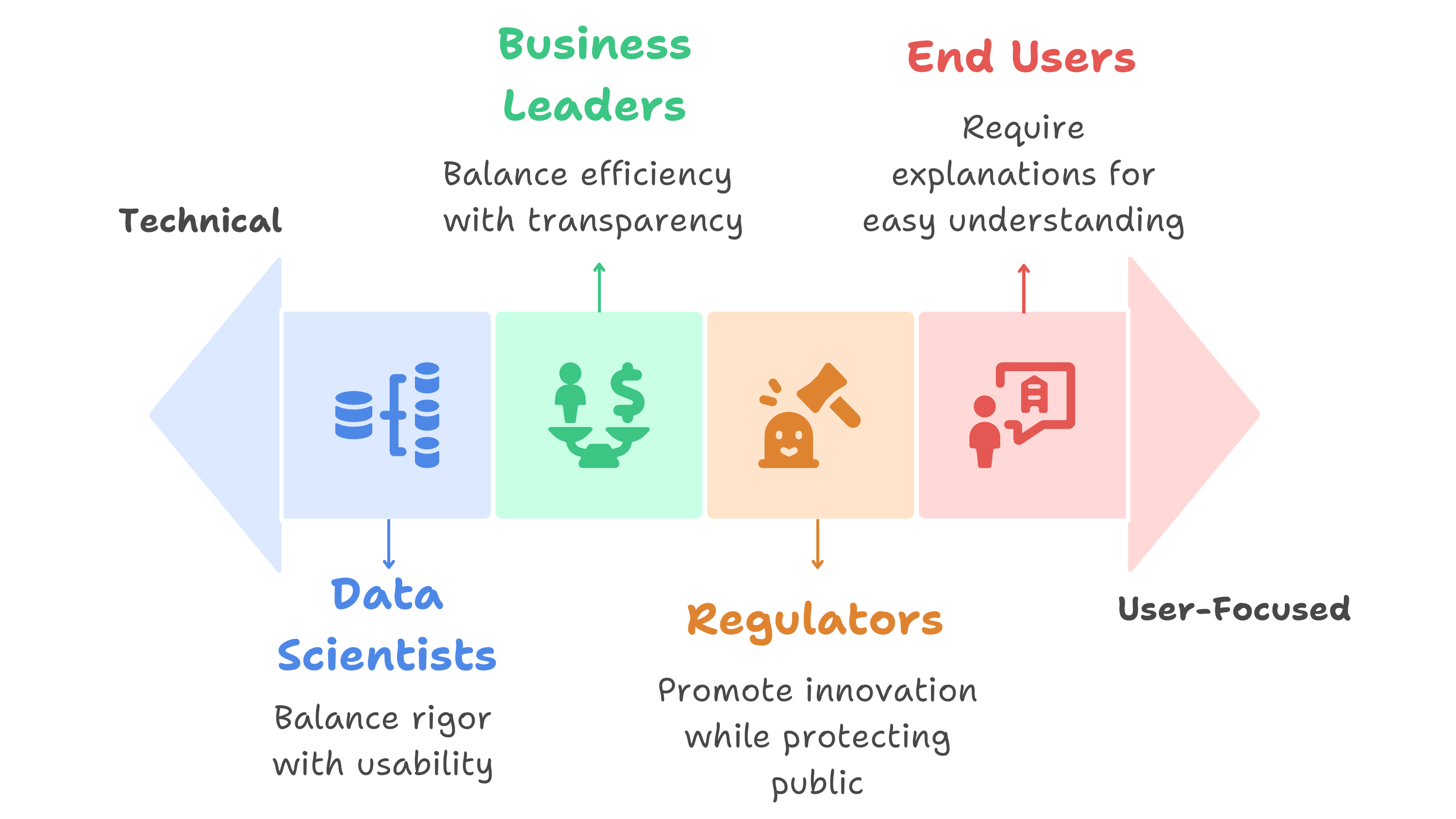}
    \caption{Interpretability Needs Across Stakeholder Groups}
    \label{fig:stakeholder_groups}
\end{figure}

Effective implementation of interpretable and explainable AI depends not only on technical solutions, but also on the collaborative efforts of diverse stakeholder groups. Each group—whether technical experts, business leaders, regulators, or end users—brings unique priorities, constraints, and responsibilities that shape the expectations and requirements for transparency (Fig.~\ref{fig:stakeholder_groups}). Recognizing and addressing these differing perspectives is essential for designing AI systems that are not only accurate but also trusted, understandable, and aligned with real-world needs. These roles are not isolated; rather, they intersect and reinforce one another in the shared pursuit of responsible, transparent AI.

\subsection{For Data Scientists and ML Specialists}
Data scientists and ML specialists serve as the primary architects of explainable  AI systems, requiring them to balance technical rigor with practical usability while remaining aware of downstream stakeholder needs across diverse application domains. These practitioners face the critical challenge of designing systems that meet both performance requirements and interpretability standards.

When developing interpretable machine learning models, practitioners should prioritize inherently interpretable architectures over post-hoc explanations whenever performance requirements allow, as these approaches provide more reliable and accurate representations of the underlying decision process. Interpretability strategies must be adapted to the specific context in which a model is applied, as domain-specific constraints and the nature of the data heavily shape what is feasible and effective~\citep{Rudin2022}. For example, rule-based models or monotonic constraints may be well-suited for financial risk scoring due to regulatory demands and structured data, whereas in natural language processing, attention mechanisms in LLMs can offer insight into how models weigh different parts of a sentence. These variations highlight the need for technical teams to align their interpretability methods with the practical and technical demands of each domain.

To ensure these interpretable systems deliver meaningful value, data scientists should establish comprehensive evaluation frameworks that extend beyond traditional accuracy metrics. These evaluation pipelines must systematically assess both technical performance and stakeholder understanding, including how explanation quality correlates with model performance indicators and prediction reliability~\citep{Jin2022}. This holistic evaluation approach ensures that interpretability efforts translate into genuine improvements in user comprehension and decision-making support rather than merely satisfying technical requirements.

\subsection{For Business Leaders and Decision Makers}

Business leaders must balance operational efficiency with transparency requirements while ensuring organizational readiness for explainable AI adoption. Not all machine learning applications require interpretability, particularly for low-stakes decisions or applications with trivial explanations and perfect reliability. We present the following guidelines which are appropriate for leaders and decision makers:

\begin{itemize}
\item \textbf{Risk Assessment}: Leaders should categorize AI applications by risk level and regulatory requirements. High-stakes applications involving safety or significant consequences require mandatory explainability protocols, as the lack of explanations during system failures can create serious problems. The task of defining which are high-stakes applications in the organization must be decided upon by technology leaders in collaboration with other relevant stakeholders. 

\item \textbf{Resource Allocation}: Time-sensitive applications require careful consideration of computational resources for explanation generation. Organizations must balance explanation quality with operational efficiency constraints \citep{Jin2022}. For example, in emergency medical diagnosis systems, radiologists must rapidly interpret critical imaging results for stroke patients where every minute affects patient outcomes. While comprehensive explanations showing detailed feature attributions across multiple image regions could provide complete understanding of the AI's decision-making process, generating such explanations creates significant delays. Organizations must therefore choose between explanation completeness and clinical workflow efficiency, often opting for faster, targeted explanations that provide sufficient insight without delaying life-critical treatment decisions.

\item \textbf{Change Management}: Users integrate AI-generated evidence into their existing decision-making processes for various downstream tasks. Organizations must prepare workflows that accommodate AI explanations within established procedures. As an example illustrating this, consider a bank loan officer who traditionally evaluates loan applications by reviewing credit scores, income statements, and employment history in a specific sequence, then makes approval decisions based on established criteria. When AI is introduced to assist with risk assessment, the officer now receives an AI recommendation along with explanations highlighting key risk factors. The organization must modify the existing loan approval workflow to include a step where officers review and interpret the AI explanation, understand how it relates to their traditional evaluation criteria, and decide whether to follow, modify, or override the AI recommendation. This requires training officers on how to incorporate AI insights into their established decision-making process and updating procedures to document how AI explanations influenced final decisions.

\item \textbf{Governance and Oversight:}
Business leaders should establish clear governance structures that define accountability for AI decisions and explanations. This includes setting up review processes for model updates and ensuring ongoing compliance with evolving regulatory requirements.

\end{itemize}

\subsection{For Regulators and Policy Makers}

Regulators face the fundamental challenge of creating frameworks that promote innovation while protecting public interests, requiring a nuanced approach that recognizes the technical realities of AI development. Research indicates that interpretability and accuracy are not necessarily in conflict when considering the complete data science process, and this balance can be adjusted based on user preferences and domain requirements. Given that interpretability needs vary significantly across domains and applications, regulatory frameworks must be flexible enough to accommodate these differences while maintaining consistent protection standards. 

To achieve meaningful transparency, regulations should prioritize end-user comprehension over technical complexity, ensuring explanations are accessible without requiring specialized expertise and align with professional decision-making patterns. This user-centered approach necessitates standardized testing protocols that assess both technical performance and real-world usability of explanations across different stakeholder groups. Furthermore, while regulations should encourage the development of inherently interpretable models as the preferred approach, they must also address the limitations of post-hoc explanations for black-box models, which can create misleading characterizations and inappropriate confidence in opaque systems, potentially undermining the very transparency goals these regulations seek to achieve.

\subsection{For End Users}

End users, whether clinicians, loan officers, or other domain experts, represent the ultimate beneficiaries of explainable AI systems and require explanations designed for easy understanding without technical knowledge in machine learning or programming. These users naturally assess explanation reasonableness as a way to evaluate AI decision quality, enabling various utilities including decision verification, trust calibration, bias identification, and knowledge discovery. For explanations to be truly effective, they must align with existing professional decision-making processes—for example, medical image interpretation involves systematic feature extraction and diagnostic reasoning that explanations should directly support rather than disrupt. Most critically, interpretable models should enable informed trust decisions rather than promoting automatic trust or distrust, as both overly trusting AI systems (known as automation bias) and users ignoring alarms (alarm fatigue) can lead to dangerous outcomes where critical errors go undetected or legitimate warnings are dismissed. Therefore, users must be properly trained to evaluate and act on AI explanations based on their quality and reliability, ensuring that transparency translates into improved decision-making outcomes rather than blind acceptance or rejection of AI recommendations. 

Recent research~\citep{han2023ignorance} reveals a critical gap in how different user groups interpret AI explanations, with laypeople basing their trust on explanation faithfulness (how accurately the explanation represents the underlying model) while domain experts rely on explanation alignment (how well explanations match their prior knowledge), suggesting that domain experts may experience cognitive biases due to their expertise. This finding is problematic because faithfulness should be the primary criterion before considering alignment with prior knowledge, yet domain experts in high-stakes domains like medicine may be trusting explanations for the wrong reasons. To address these cognitive biases and interpretation gaps, organizations must implement comprehensive training programs that help all end users understand both the capabilities and limitations of AI explanations, including developing skills for recognizing when explanations may be misleading, when additional verification is needed, and how to properly evaluate explanation faithfulness regardless of their domain expertise.

\section{Incorporating Interpretability into AI Development Workflow}

\begin{figure}[H]
    \centering
    \includegraphics[scale=0.3]{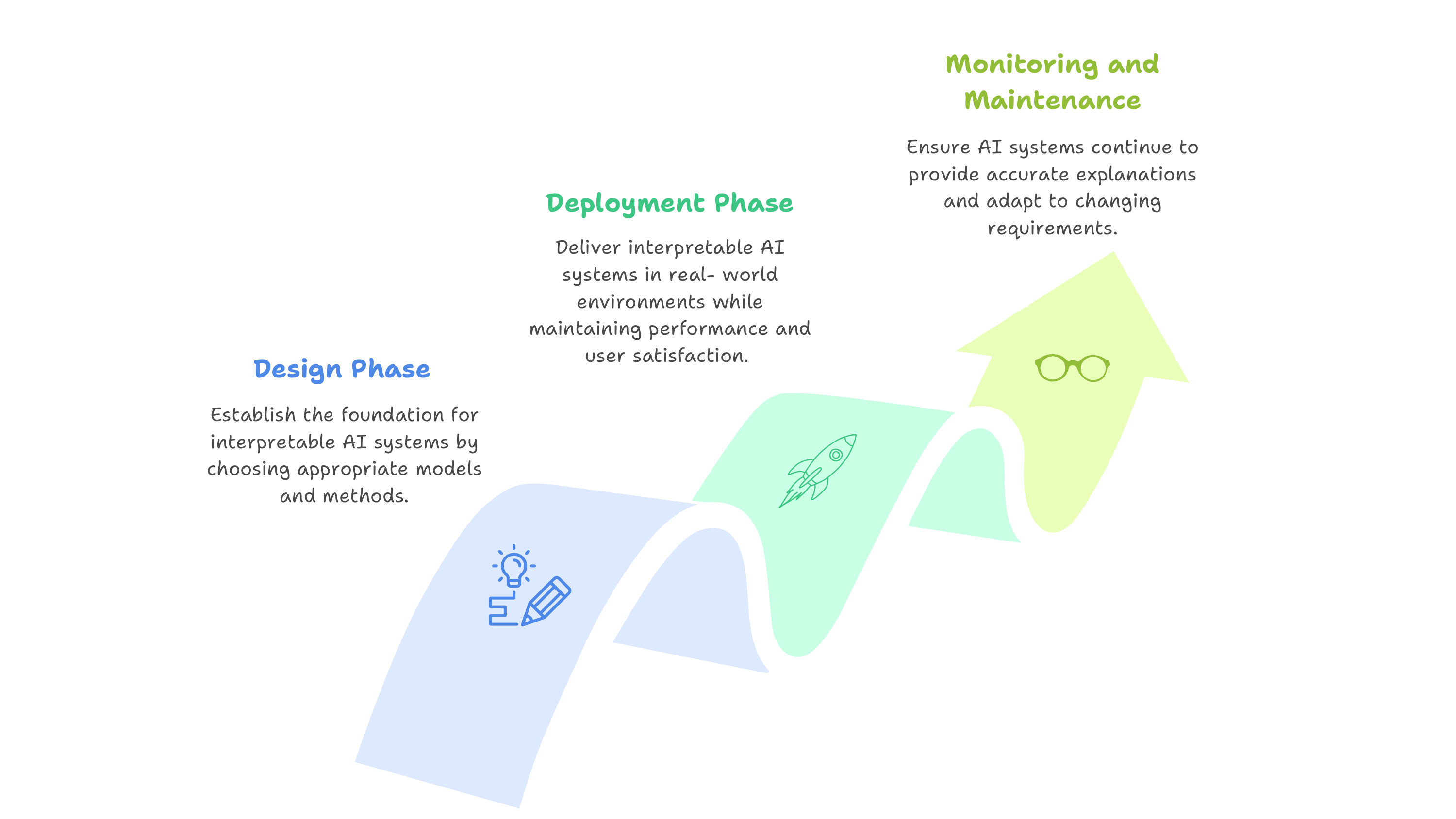}
    \caption{Integrating Interpretability Across the AI Lifecycle}
    \label{fig:ai_lifecycle}
\end{figure}

The integration of interpretability into AI development requires a systematic approach that spans the entire machine learning lifecycle. Rather than treating explainability as an afterthought, organizations must embed interpretability considerations from initial design through ongoing maintenance (Fig.~\ref{fig:ai_lifecycle}).

\subsection{Design Phase: Choosing Appropriate Models and Methods}

The design phase establishes the foundation for interpretable AI systems by making informed decisions about model architecture, data pre-processing, and explanation methods based on stakeholder requirements and domain constraints.

\subsubsection{Problem Formulation and Requirements Analysis}
The first step involves defining interpretability requirements based on the application domain and stakeholder needs. Interpretable models incorporate domain-specific constraints that facilitate human understanding, and these constraints vary significantly across different applications. Organizations should conduct thorough stakeholder analysis to understand who will interact with the AI system and what types of explanations they require.

For high-stakes applications, the choice between inherently interpretable models and post-hoc explanation methods becomes critical. Low-stakes decisions may not require interpretability, but applications involving human safety, legal compliance, or medical decisions should prioritize inherently interpretable models.

\subsubsection{Model Architecture Selection}
The choice of model architecture significantly impacts interpretability potential. Different data types require different interpretability approaches; sparse representations work well for tabular data, while computer vision applications benefit from specialized interpretable neural network architectures. Different model types offer varying levels of inherent interpretability, each with distinct advantages and limitations across application domains. Linear models provide direct coefficient interpretation that works well for tabular data and simple NLP tasks but may lack expressiveness for complex relationships found in image analysis or natural language understanding. Tree-based models offer rule-based explanations that mirror human decision processes, making them particularly effective for structured data applications and some object recognition tasks where decision boundaries can be clearly articulated. Ensemble methods balance predictive power with interpretability through feature importance measures, proving valuable across diverse applications from text classification to medical diagnosis. Neural networks, while powerful for complex tasks like computer vision, advanced NLP, and language generation, typically require specialized architectures or post-hoc methods for interpretability, with techniques ranging from attention mechanisms in transformers for language processing to convolutional layer visualization for image analysis. The choice of model type should therefore align with both the complexity requirements of the specific application domain and the interpretability needs of the end users, whether they are analyzing financial data, interpreting medical images, or understanding automated text generation.

\subsubsection{Data Preparation Strategy}
Data preprocessing decisions significantly impact subsequent interpretability. Feature engineering should prioritize meaningful, domain-relevant variables that align with stakeholder mental models. The creation of interpretable features often requires domain expertise and close collaboration with end users to ensure explanations will be comprehensible and actionable.

\subsection{Deployment Phase: Explaining AI in Production}

The deployment phase addresses the practical challenges of delivering interpretable AI systems in real-world environments while maintaining performance, reliability, and user satisfaction. Successful production deployment requires careful orchestration of technical architecture, user interface design, and quality assurance processes that collectively ensure explanation systems can operate effectively under operational constraints.

Production architecture design must balance explanation quality with stringent system performance requirements, particularly in time-sensitive applications where computational resources and latency constraints significantly impact explanation generation capabilities. Effective caching strategies become essential, involving pre-computation of explanations for common scenarios to minimize real-time processing overhead and reduce response latency. This approach requires sophisticated prediction of likely explanation requests and intelligent storage management that can quickly retrieve relevant pre-computed explanations while maintaining freshness and accuracy.

Adaptive complexity mechanisms provide another crucial architectural component, enabling systems to deliver different explanation depths based on user expertise levels, available time constraints, and specific decision contexts. Novice users may require comprehensive, step-by-step explanations with extensive contextual information, while expert users might prefer concise summaries that highlight only the most critical decision factors. This adaptability must be implemented through scalable patterns that ensure explanation systems can handle production-level traffic without degrading performance, incorporating load balancing, resource allocation strategies, and efficient processing pipelines that can scale horizontally as demand increases.

Robust fallback mechanisms represent a critical safety consideration, maintaining core system functionality even when explanation generation encounters failures or experiences unexpected delays. These mechanisms might include simplified explanation alternatives, cached historical explanations for similar cases, or graceful degradation that allows the AI system to continue operating with reduced interpretability rather than complete failure.

Seamless user interface integration ensures that explanations enhance rather than disrupt existing workflows, requiring deep understanding of how users naturally inspect AI-generated evidence and incorporate it into their decision-making processes for various downstream tasks. Contextual presentation strategies display explanations at optimal points in user workflows, providing relevant information precisely when users need it without overwhelming them with unnecessary detail or interrupting critical decision processes.

Progressive disclosure techniques allow users to access different levels of explanation detail according to their immediate needs and available cognitive resources. Initial presentations might provide high-level summaries with options to drill down into specific aspects of the decision process, enabling users to control the depth of information they receive based on their confidence levels, time constraints, and expertise. Multi-modal communication approaches combine visual representations, textual descriptions, and interactive explanation formats to accommodate different learning styles and decision-making preferences, while customization options enable users to adjust explanation preferences based on their domain expertise, role requirements, and personal preferences.

Comprehensive quality assurance and testing protocols ensure explanation systems maintain reliability and accuracy under realistic operational conditions. This encompasses stress testing explanation generation capabilities under high computational loads to identify performance bottlenecks and ensure graceful degradation under extreme conditions. Validation procedures verify explanation consistency across different system configurations, deployment environments, and user access patterns, ensuring that explanations remain stable and reliable regardless of the specific technical infrastructure or user interaction modalities.

Continuous verification processes become particularly critical as underlying AI models undergo updates, retraining, or architectural modifications, requiring systematic testing to ensure that explanations remain accurate representations of updated model behavior. This ongoing quality assurance must account for the dynamic nature of production AI systems, where model performance, data distributions, and user requirements may evolve over time, demanding explanation systems that can adapt while maintaining their interpretive fidelity and user utility.

\subsection{Monitoring and Maintenance}

The monitoring and maintenance phase ensures that interpretable AI systems continue to provide accurate, useful explanations throughout their operational lifecycle while adapting to changing requirements and conditions.

\textbf{Explanation Quality Monitoring:}
Continuous monitoring systems should track explanation quality metrics to detect degradation or inconsistencies. This includes automated checks for explanation stability, user feedback analysis, and periodic validation against ground truth when available. Organizations should establish thresholds for explanation quality metrics and implement alerting systems when these thresholds are exceeded.

Key monitoring activities include:
\begin{itemize}
\item \textbf{Drift Detection}: Monitor for changes in explanation patterns that may indicate model or data drift
\item \textbf{User Engagement Tracking}: Analyze how users interact with explanations to identify improvement opportunities
\item \textbf{Performance Impact Assessment}: Continuously evaluate the computational cost of explanation generation
\item \textbf{Error Rate Monitoring}: Track instances where explanations may have misled users or contributed to incorrect decisions
\end{itemize}

\textbf{Adaptive Improvement Processes:}
Interpretable AI systems should incorporate feedback mechanisms that enable continuous improvement of explanation quality and relevance. This includes collecting user feedback on explanation usefulness, analyzing patterns in user behavior with explanations, and updating explanation methods based on new research and best practices.

\textbf{Regulatory Compliance Maintenance:}
As regulatory requirements evolve, organizations must ensure their interpretable AI systems remain compliant. This requires ongoing assessment of explanation adequacy against current regulations, documentation of explanation methodologies for audit purposes, and implementation of processes for updating explanation systems to meet new requirements.

The maintenance phase should also include regular reviews of interpretability requirements with stakeholders, as user needs and domain understanding may evolve over time. Organizations should establish processes for systematically updating explanation methods and retraining models while maintaining explanation quality and consistency.


\section{Standardized Reporting Framework for Interpretable Models}

\subsection{Proposed Reporting Template}

To ensure consistent documentation and evaluation of interpretable AI systems across organizations and domains, we propose a standardized reporting template that captures essential elements of model interpretability. This template serves as a comprehensive guide for practitioners to document their interpretable AI implementations, enabling better transparency, reproducibility, and compliance with regulatory requirements.

\begin{figure}[H]
    \centering
    \includegraphics[scale=0.3]{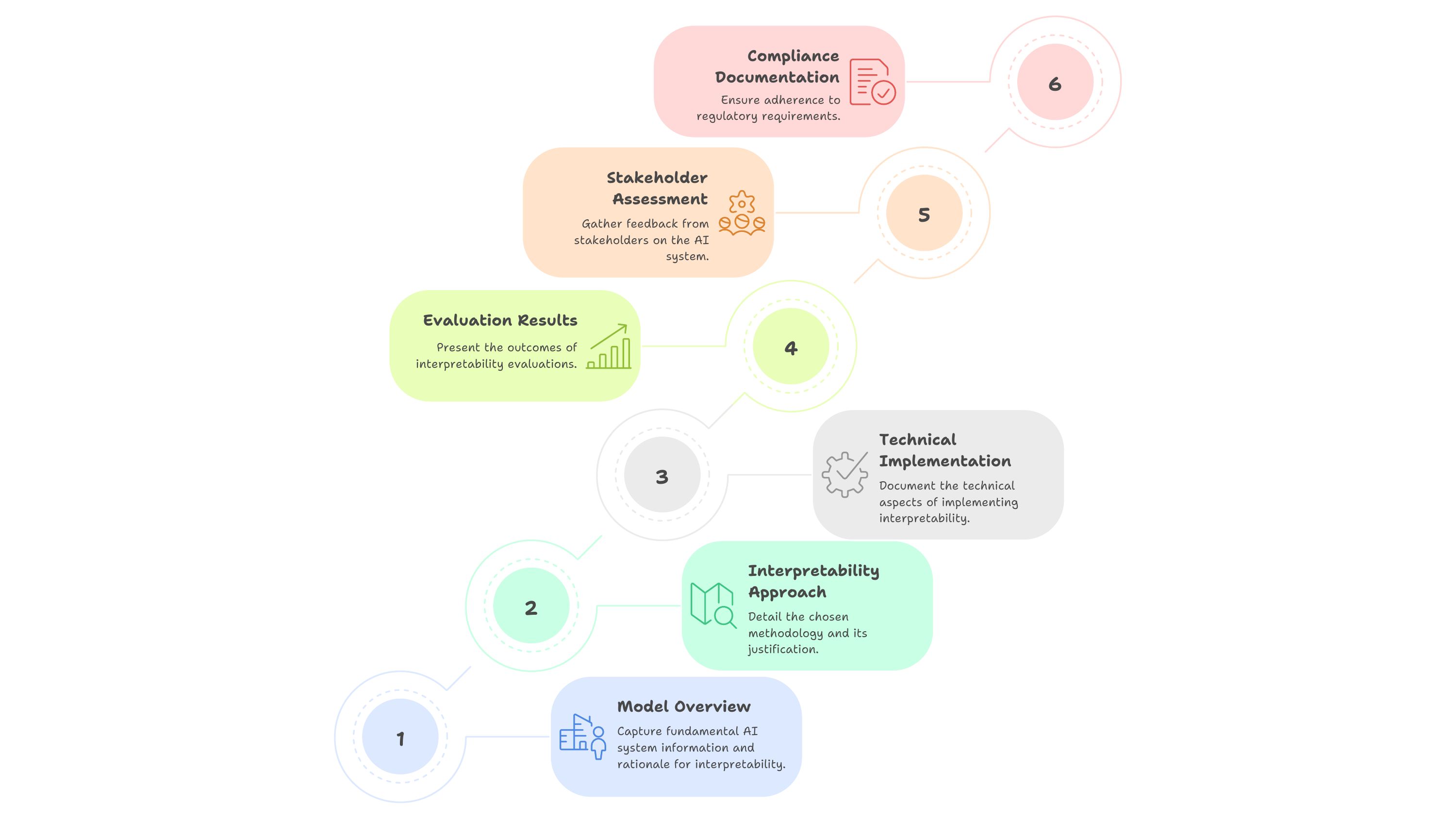}
    \caption{Key Components of a Standardized Reporting Template for Interpretable AI Systems}
    \label{fig:reporting_structure}
\end{figure}

The proposed template consists of six major sections: Model Overview, Interpretability Approach, Technical Implementation, Evaluation Results, Stakeholder Assessment, and Compliance Documentation (Fig.~\ref{fig:reporting_structure}). Each section contains specific fields and requirements designed to capture the most critical aspects of interpretable AI systems while remaining flexible enough to accommodate diverse application domains and organizational contexts.

\textbf{Section 1: Model Overview}
This section captures fundamental information about the AI system, including the problem domain, intended use case, target users, and business context. It should specify the model architecture, data sources, training methodology, and performance characteristics. Additionally, it documents the rationale for requiring interpretability in this specific application and the expected benefits for stakeholders.

\textbf{Section 2: Interpretability Approach}
This section details the chosen interpretability methodology, whether inherently interpretable models or post-hoc explanation techniques. It should justify the selection of specific interpretability approaches based on domain requirements, user needs, and technical constraints. The documentation should include a clear mapping between interpretability techniques and their intended purposes, such as global model understanding, local prediction explanations, or feature importance analysis.

\subsection{Essential Elements to Document}

The standardized reporting framework requires documentation of twelve essential elements that collectively provide a comprehensive view of the interpretable AI system's design, implementation, and effectiveness.

\textbf{Interpretability Requirements and Justification:} Document the specific reasons why interpretability is necessary for this application, including regulatory requirements, stakeholder needs, risk assessment outcomes, and business objectives. This section should clearly articulate the expected benefits of interpretability and how they align with organizational goals.

\textbf{Stakeholder Analysis:} Identify all stakeholders who will interact with or be affected by the AI system's explanations, including their technical background, domain expertise, decision-making responsibilities, and specific information needs. This analysis should inform the design of explanation interfaces and content.

\textbf{Model Architecture and Design Decisions:} Provide detailed documentation of the chosen model architecture, including the rationale for selecting inherently interpretable models versus complex models with post-hoc explanations. Document any architectural modifications made to enhance interpretability and their impact on model performance.

\textbf{Explanation Generation Methods:} Describe the specific techniques used to generate explanations, including their theoretical foundations, implementation details, computational requirements, and limitations. For post-hoc methods, document validation procedures used to ensure explanation fidelity and reliability.

\textbf{Technical Performance Metrics:} Report both predictive performance metrics (accuracy, precision, recall, etc.) and interpretability-specific metrics (fidelity, stability, consistency, completeness). Include statistical significance tests and confidence intervals where appropriate, and acknowledge any trade-offs between performance and interpretability.

\textbf{Human-Centered Evaluation Results:} Document the results of user studies, including task performance measurements, trust calibration assessments, mental model alignment evaluations, and actionability assessments. Report both quantitative metrics and qualitative feedback from actual end users.

\textbf{Deployment Architecture:} Describe the technical infrastructure used to deliver explanations in production, including caching strategies, performance optimization techniques, fallback mechanisms, and scalability considerations. Document any limitations or constraints that affect explanation availability or quality.

\textbf{Quality Assurance Procedures:} Detail the processes used to ensure explanation quality, including validation protocols, monitoring systems, alerting mechanisms, and continuous improvement procedures. Document quality thresholds and the actions taken when these thresholds are exceeded.

\textbf{Regulatory Compliance:} Provide evidence of compliance with relevant regulations and standards, including documentation of explanation adequacy, audit trails, and processes for maintaining compliance as requirements evolve. Include references to specific regulatory frameworks and their requirements.

\textbf{Risk Assessment and Mitigation:} Document potential risks associated with the interpretable AI system, including risks of misleading explanations, over-reliance on AI recommendations, or misinterpretation of explanations. Describe mitigation strategies and monitoring procedures for each identified risk.

\textbf{Maintenance and Update Procedures:} Outline processes for maintaining explanation quality over time, including procedures for updating explanation methods, retraining models, monitoring explanation drift, and incorporating user feedback. Document version control procedures and change management processes.

\textbf{Limitations and Known Issues:} Acknowledge limitations of the interpretability approach, including scenarios where explanations may be incomplete or misleading, technical constraints that affect explanation quality, and areas where further research or development is needed.

\subsection{Case Study: Implementing the Reporting Framework}

To demonstrate the practical application of the standardized reporting framework, we present a case study of its implementation in a healthcare AI system for diagnostic imaging. This case study illustrates how the framework can be adapted to specific domain requirements while maintaining consistency and comprehensiveness.

\begin{subsection}{Medical Imaging Diagnostic Support System}

\textbf{Model Overview:} The system assists radiologists in detecting lung cancer in chest X-rays, serving as a second opinion tool in a hospital radiology department. The target users are board-certified radiologists with varying levels of experience in chest imaging. The business context involves improving diagnostic accuracy while maintaining radiologist autonomy in final decision-making.

\textbf{Interpretability Approach:} Given the high-stakes medical context and regulatory requirements, the team selected an inherently interpretable ensemble model combining generalized additive models with attention-based neural networks. This hybrid approach enables both global feature understanding and local explanation of specific predictions while maintaining competitive diagnostic accuracy.

\textbf{Technical Implementation:} The model processes chest X-rays through a specialized interpretable neural network architecture that generates pixel-level attribution maps highlighting regions most relevant to the cancer detection decision. Global explanations show the model's overall decision patterns across different demographic groups and imaging conditions.

\textbf{Evaluation Results:} The system achieved 94.2

\textbf{Regulatory Compliance:} The system meets FDA requirements for AI/ML-based medical devices, with comprehensive documentation of explanation methodologies and validation procedures. Regular audits ensure ongoing compliance with evolving regulations.

\end{subsection}

This case study demonstrates how the reporting framework captures essential information while remaining practical for real-world implementation. The standardized format enables comparison across different systems and domains while ensuring that critical aspects of interpretable AI are consistently documented and evaluated.

\section{Evaluation Methods for Model Interpretability and Explainability}

The evaluation of explainable artificial intelligence (XAI) methods presents unique challenges as it requires measuring subjective qualities like comprehensibility, faithfulness, and usefulness rather than traditional predictive performance metrics. Quantitative assessment methods provide objective measures through functionally-grounded evaluation, including fidelity measures that test how faithfully explanations represent model behavior through perturbation-based approaches, stability measures that assess explanation robustness across similar inputs, and completeness metrics that evaluate whether explanations capture the full scope of model behaviour \citep{doshi2017considerations, samek2017evaluating, adebayo2018sanity}. For specialized domains like time-series data, novel metrics such as area under the top curve and modified F1 scores have been developed to address temporal dependencies and sequential relationships \citep{turbe2023evaluation}.

Human-centered evaluation directly assesses explanation quality from the user perspective, recognizing that interpretability ultimately depends on human comprehension and utility. This approach follows a three-level framework: application-grounded evaluation with real end-users performing actual domain tasks, human-grounded evaluation using simplified tasks with lay participants, and functionally-grounded evaluation using computational proxies. Task-based evaluations include forward simulation where participants predict model outputs from explanations, counterfactual reasoning to test understanding of decision boundaries, and trust assessment to measure appropriate calibration of user confidence in model predictions \citep{doshi2017considerations, nauta2023systematic}.

A comprehensive evaluation framework organizes explanation quality into twelve properties (Co-12) across content dimensions (correctness, completeness, consistency), presentation aspects (compactness, composition, confidence), and user considerations (context, coherence, controllability). The selection of evaluation methods should align with application context, employing multiple quantitative metrics for general-purpose assessment, prioritizing human-centered evaluation with domain experts for specific applications, and using functionally-grounded methods for iterative research and development \citep{nauta2023systematic, velmurugan2024guidelines}.

Comprehensive evaluation reporting should include:

\textbf{Methodology Transparency:} Clearly describe evaluation procedures, including data preprocessing, perturbation strategies, and baseline comparisons. Report both positive and negative results to avoid publication bias.

\textbf{Multi-Dimensional Assessment:} Report metrics across multiple Co-12 properties rather than focusing on single measures. Acknowledge trade-offs between different quality aspects.

\textbf{Statistical Rigor:} Include confidence intervals, significance tests, and multiple random seeds when applicable. Report both individual instance results and aggregate statistics.

\textbf{Reproducibility Information:} Provide sufficient implementation details, dataset descriptions, and code availability to enable replication studies.

\textbf{Limitation Discussion:} Explicitly acknowledge evaluation limitations, including cases where methods may not generalize or where evaluation metrics may not capture all relevant quality aspects.

The field continues evolving toward more standardized evaluation practices, with growing recognition that comprehensive assessment requires combining multiple evaluation approaches rather than relying on single metrics or methodologies.

\begin{figure}[H]
    \centering
    \includegraphics[width=1.0\textwidth]{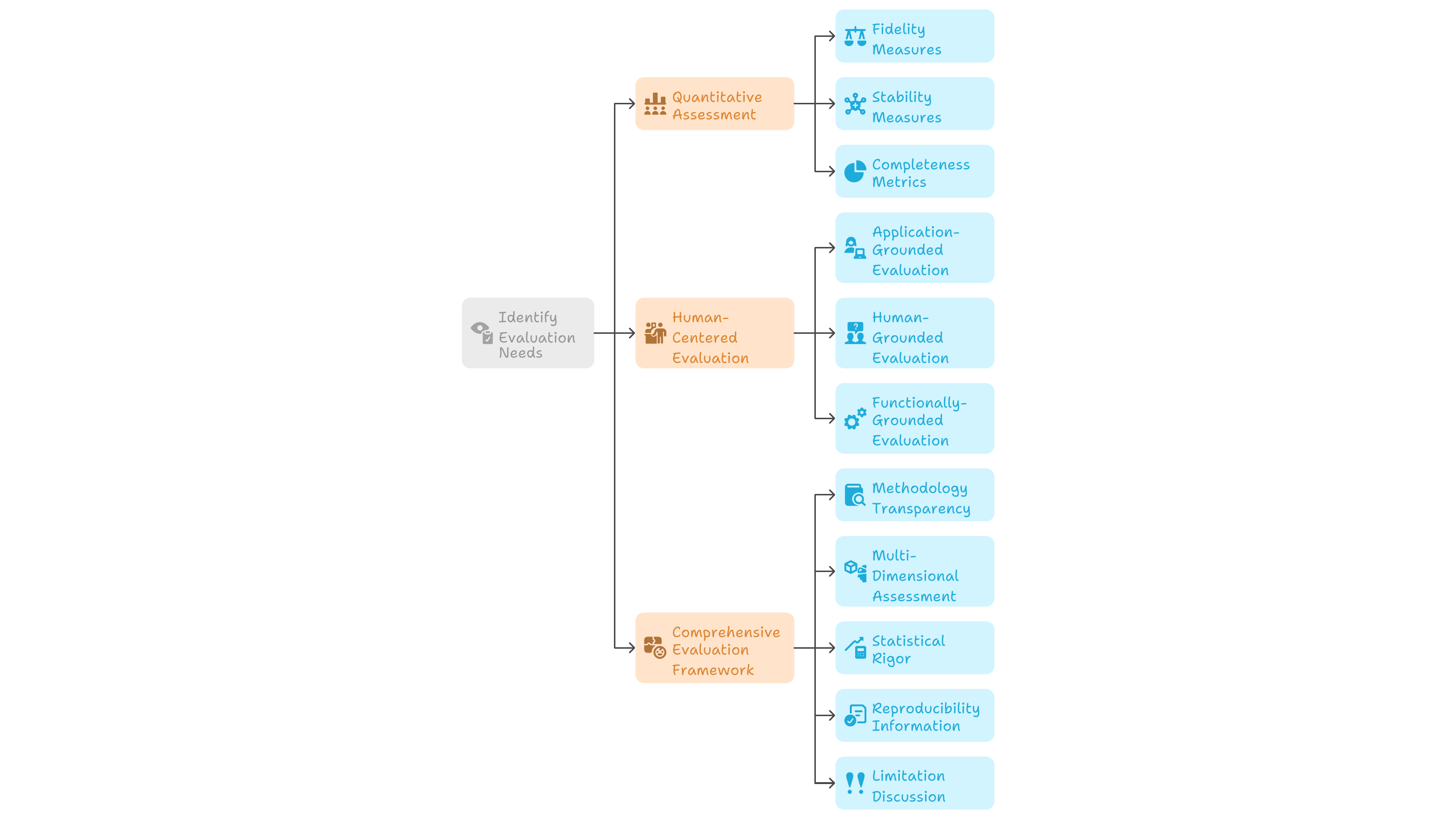}
    \caption{Multi-Level Evaluation Framework for Interpretability Methods}
    \label{fig:evaluation_methods}
\end{figure}

\section{Challenges and Limitations}
\begin{figure}[H]
    \centering
    \includegraphics[scale=0.3]{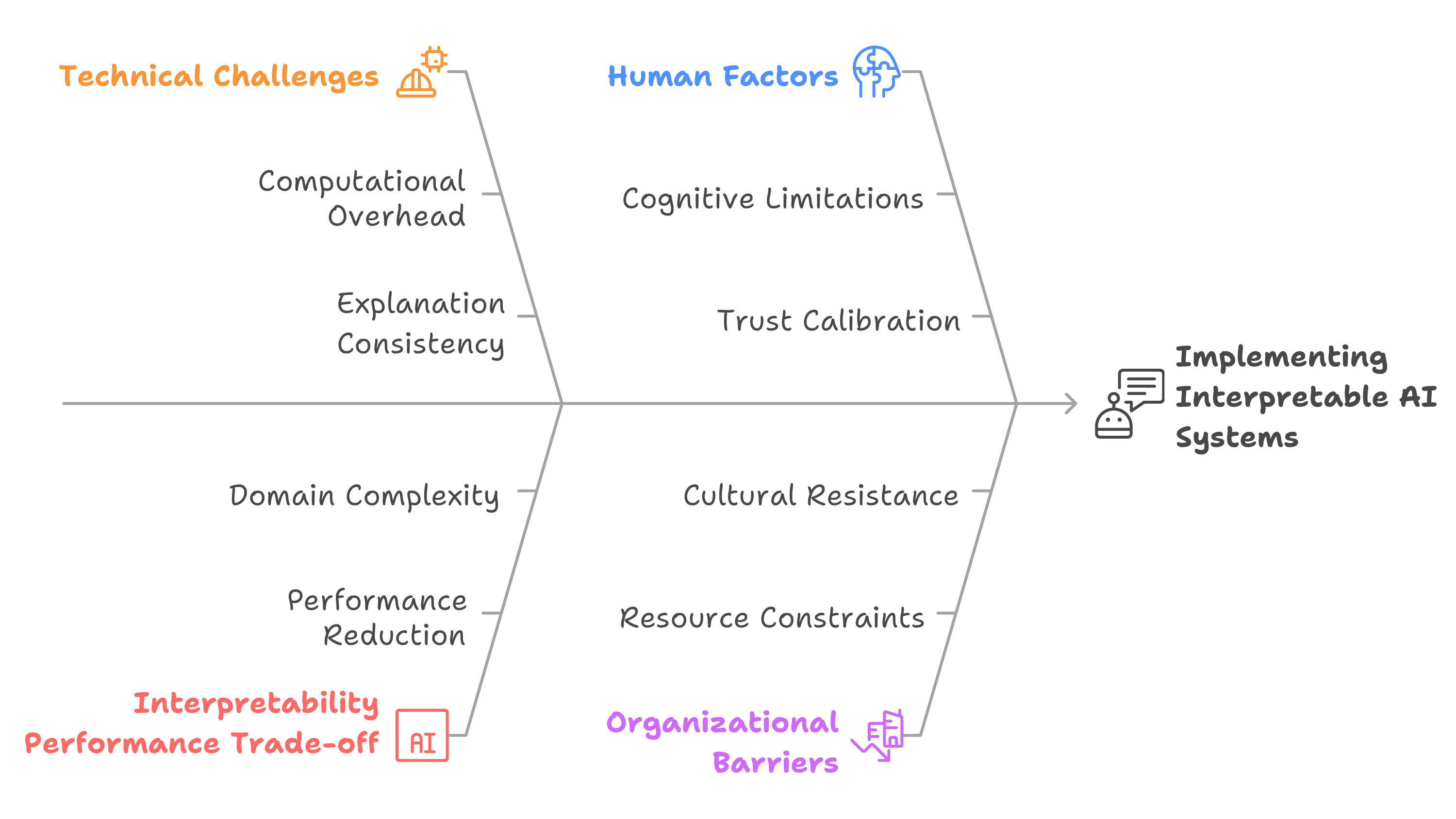}
    \caption{Key Challenges to Implementing Interpretable AI Systems}
    \label{fig:challenges}
\end{figure}

Despite growing interest and advancements in interpretable AI, several technical, human, and organizational barriers continue to impede its widespread implementation. These challenges are often interdependent and must be addressed holistically to build AI systems that are not only accurate but also trustworthy, transparent, and aligned with human values. Key challenges across four dimensions emerge: technical limitations, trade-offs with performance, human cognitive factors, and institutional constraints—that impact the effective integration of interpretability into AI systems(Fig.~\ref{fig:challenges}).

\subsection{Technical Challenges}

The implementation of interpretable AI systems faces several significant technical challenges that organizations must navigate carefully. Computational overhead represents one of the most immediate concerns, as generating high-quality explanations often requires substantial processing power and time, particularly for complex models and large datasets. Real-time applications face especially acute constraints, where the latency introduced by explanation generation can compromise system responsiveness and user experience.

Explanation consistency poses another critical challenge, particularly when multiple interpretation methods are applied to the same model or decision. Different explanation techniques may yield conflicting insights about the same prediction, creating confusion for users and undermining confidence in the interpretability system. This inconsistency can arise from fundamental differences in how various methods operate, from gradient-based approaches that may be sensitive to input perturbations to sampling-based methods that introduce stochastic variation.

Scalability challenges emerge as organizations attempt to deploy interpretable AI systems across large-scale operations. Explanation methods that work well for individual predictions or small datasets may not scale effectively to enterprise-level deployments with millions of daily predictions. The infrastructure required to support explanation generation, storage, and delivery can become prohibitively expensive, particularly for organizations operating under tight budget constraints.

\subsection{The Interpretability-Performance Trade-off}

A common perception is that increasing model interpretability comes at the cost of performance. However, recent arguments suggest this is a false dichotomy. Rudin et al. (2022) advocate that in many high-stakes applications, interpretable models can match or even outperform black-box models.

The interpretability-performance trade-off manifests differently across application domains and use cases. In domains where the underlying relationships are relatively simple or well-understood, inherently interpretable models like linear regression or decision trees can achieve performance comparable to more complex alternatives while providing complete transparency. However, in domains involving high-dimensional data, complex non-linear relationships, or subtle pattern recognition tasks, the performance gap between interpretable and black-box models may be more significant.

Organizations must carefully assess this trade-off in the context of their specific requirements. For high-stakes decisions where understanding the reasoning process is paramount—such as medical diagnosis, loan approval, or criminal justice applications—accepting some performance reduction in favor of interpretability may be justified. Conversely, for applications where performance is critical and the consequences of individual errors are manageable, organizations might prioritize accuracy over interpretability.

Use of inherently interpretable models is especially crucial in domains such as medicine, where understanding and accountability are paramount. For lower-risk applications (e.g., targeted advertising), black-box models may be acceptable provided proper governance and monitoring are in place.

\subsection{Human Factors and Cognitive Limitations}

The effectiveness of interpretable AI systems is fundamentally limited by human cognitive capacity and information processing constraints. Users have finite attention spans and working memory, meaning that explanations must be carefully designed to convey the most important information without overwhelming cognitive resources. Complex explanations with too much detail can be counterproductive, leading to information overload and poor decision-making.

Cognitive biases present another significant challenge, as users may interpret explanations through the lens of their existing beliefs and expectations. Confirmation bias can lead users to focus on explanation elements that support their preconceptions while ignoring contradictory information. Anchoring bias may cause users to over-weight the first piece of explanation information they encounter, while availability bias might lead them to overestimate the importance of easily recalled examples or patterns.

Domain expertise creates additional complexity, as expert users and novices have fundamentally different information needs and interpretation patterns. Experts may prefer concise, technical explanations that align with their professional knowledge, while novices require more comprehensive, educational explanations that build understanding from basic principles. Designing explanation systems that effectively serve both audiences without compromising utility for either represents a significant challenge.

Trust calibration presents a particularly complex human factor challenge. Users must develop appropriate levels of trust in AI systems—neither over-trusting nor under-trusting the technology. Explanations play a crucial role in this calibration process, but poorly designed explanations can lead to inappropriate trust levels, whether excessive confidence in flawed systems or unnecessary skepticism toward reliable ones.

\subsection{Organizational Barriers to Adoption}

Organizations face numerous structural and cultural barriers to implementing interpretable AI systems effectively. Resource constraints often limit the ability to invest in the additional development time, computational infrastructure, and specialized expertise required for interpretable AI. Organizations operating under tight budgets or aggressive timelines may view interpretability as a luxury rather than a necessity, particularly when regulatory requirements are unclear or loosely enforced.

Cultural resistance to transparency can present significant obstacles, particularly in organizations with hierarchical decision-making structures or competitive cultures where information is viewed as power. Some stakeholders may resist interpretable AI systems because they fear that transparency will expose inefficiencies, biases, or errors in existing processes. This resistance can manifest as reluctance to provide training data, participate in user studies, or adopt new workflows that incorporate AI explanations.

Technical debt and legacy system integration pose practical implementation challenges. Organizations with established AI systems may find it difficult to retrofit interpretability capabilities without significant architectural changes. The cost and risk associated with modifying production systems can be substantial, particularly for organizations in regulated industries where system changes require extensive validation and approval processes.

Skills gaps present another significant barrier, as interpretable AI requires expertise that spans multiple disciplines including machine learning, user experience design, domain knowledge, and regulatory compliance. Organizations may struggle to find individuals with the necessary combination of skills or to develop such expertise internally. This challenge is compounded by the rapidly evolving nature of interpretability research and the lack of standardized training programs.

Change management represents a critical organizational challenge, as implementing interpretable AI often requires modifications to existing workflows, decision-making processes, and organizational roles. Users may resist adopting new tools or processes, particularly if they perceive the changes as threatening their autonomy or expertise. Successful implementation requires careful change management strategies that address user concerns, provide adequate training, and demonstrate clear benefits.

\section{Interpretability Applications and Cross-Industry Learnings}
To better understand how interpretability and explainability function in real-world applications, we analyzed case studies spanning sectors such as finance, healthcare, telecommunications, infrastructure, and human resources. These applied efforts offered valuable insights into both technical and organizational dynamics that shape the successful adoption of interpretable machine learning (ML) systems.

\subsection{Interpretability Across Domains: Purpose and Practice}
As AI systems continue to permeate a variety of sectors, from finance and healthcare to infrastructure, education, and human resources, there is a growing recognition that interpretability is not merely a desirable feature but a necessary one. The role of interpretability varies significantly depending on the domain, yet a common thread emerges: the need to understand, trust, and justify machine learning outputs in ways that align with business goals, compliance demands, and end-user expectations.

In high-stakes domains like healthcare or finance, interpretability is often motivated by regulatory scrutiny and ethical accountability. In technical domains like infrastructure or telecommunications, it supports root cause analysis and system debugging. Meanwhile, in human-centered fields such as HR and education, interpretability facilitates fairness, transparency, and improved stakeholder communication.

Interpretability applications in these diverse settings reveals that successful interpretability efforts tend to share key characteristics of deliberate planning, alignment with stakeholder needs, and integration into the broader system development process. Regardless of sector, teams that treated interpretability as a core design requirement, rather than a post-hoc addition, were better positioned to meet both technical and organizational objectives.

\subsection{Key Cross-Sector Observations and Insights}
\begin{figure}[H]
    \centering
    \includegraphics[scale=0.3]{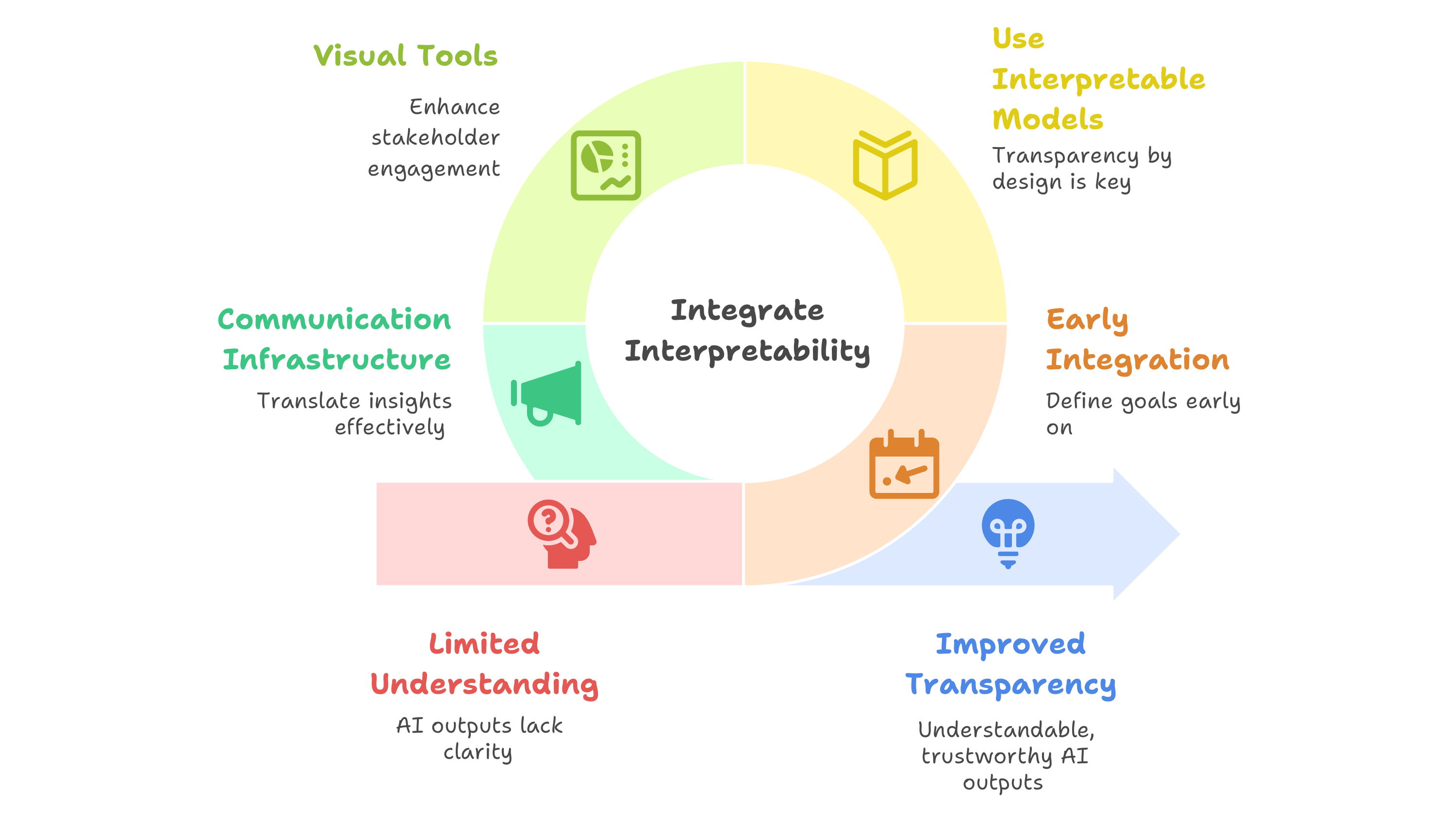}
    \caption{Adoption of AI Interpretability Across Domains}
    \label{fig:integration_lifecycle}
\end{figure}
Several recurring themes emerged across the case studies (Fig.~\ref{fig:integration_lifecycle}):

\begin{itemize}

\item \textbf {Early Integration of Interpretability is Critical}:
Projects that defined interpretability goals early in development were better able to align models, explanation techniques, and stakeholder needs. Delayed consideration often led to costly retrofitting and limited effectiveness.

\item \textbf {Inherently Interpretable Models Are Often Better Aligned with Stakeholders}:
Models like Explainable Boosting Machines (EBMs), Generalized Additive Models (GAMs), and Neural Additive Models (NAMs) offered transparency by design and were easier for non-technical stakeholders to understand. These were particularly effective in domains requiring high accountability.

\item \textbf {Trade-offs Between Accuracy and Interpretability Are Contextual}:
In domains like fraud detection and clinical risk prediction, interpretable models occasionally underperformed black-box counterparts but offered significant advantages in trust and compliance. Teams often had to balance predictive performance with clarity and audit-ability.

\item \textbf {Limitations of Post-hoc Methods in Complex Use Cases}:
Tools like SHAP and LIME were widely used but often struggled with stability, sensitivity to hyperparameters, and inconsistent outputs when applied to large language models or high-dimensional data. These limitations highlight the need for careful method selection based on domain and model architecture.

\item \textbf {Explainability for Generative Models is Still Maturing}:
Use cases involving large language models (e.g., LLaMA, Falcon, Mistral) revealed gaps in current interpretability tools. Techniques such as attribution mapping and perturbation helped expose token relationships but were constrained by the models' complexity and opacity.

\item \textbf {Visual Tools and Communication Infrastructure Enhance Stakeholder Engagement}:
Projects that included dashboards, interactive plots, or structured reporting templates were more effective at translating technical insights into actionable understanding for business users, auditors, or compliance teams.

\item \textbf {Interpretability is a Socio-Technical Challenge}:
Success was often driven more by process, communication, and alignment than technical execution alone. Teams that fostered feedback loops, included domain experts early, and regularly reflected on their approach achieved more sustainable outcomes.
\end{itemize}

\subsection{Domain-Specific Highlights}
Case studies across diverse industries including finance, telecommunications, healthcare, human resources, infrastructure, and education demonstrate the varied ways in which interpretability techniques can be applied to domain-specific machine learning challenges. These applications surfaced key technical and organizational considerations, highlighting the trade-offs, limitations, and opportunities inherent in building transparent and trustworthy AI systems. Despite sector-specific complexities, the underlying techniques proved adaptable and effective across a wide range of use cases, reinforcing their relevance to any industry pursuing responsible AI adoption.

\begin{itemize}
 \item \textbf{Finance and Fraud Detection:} In domains with high regulatory oversight, such as fraud detection and credit scoring, techniques like SHAP, LIME, and Explainable Boosting Machines (EBMs) were used to enhance transparency. Post-hoc tools such as SHAP provided fine-grained insight into feature importance, but frequently exhibited instability when applied to complex or imbalanced datasets. In contrast, inherently interpretable models like EBMs produced more stable, rule-based outputs that resonated better with non-technical stakeholders—even when sacrificing some predictive performance or sensitivity to rare events.

 \item \textbf{Healthcare:} Clinical applications such as patient outcome prediction and hospital readmission risk modeling leaned on techniques like Generalized Additive Models (GAMs) and Partial Dependence Plots (PDPs) to deliver transparent decision logic. These approaches proved particularly valuable in fostering clinician trust in model outputs. However, explainability was harder to maintain in contexts involving sparse or high-dimensional data, sometimes requiring simplified modeling approaches or hybrid architectures to ensure interpretability without compromising utility.

 \item \textbf{LLMs and Document AI:} Use cases such as resume screening, banking chatbot response generation, and optical character recognition (OCR) pipelines applied interpretability tools to large-scale generative models like LLaMA, Falcon, and Mistral. Token-level attribution and perturbation techniques provided partial insight into model reasoning, but their effectiveness was limited by the opaque nature of attention-based architectures. These challenges point to an urgent need for domain-specific interpretability methods tailored to generative and sequence-based models, particularly in compliance-sensitive fields like HR and financial services.

 \item \textbf{Infrastructure and Industrial Systems:} For tasks such as anomaly detection in sensor and telemetry data, glass-box models like EBMs and GAMs offered valuable interpretability benefits over black-box methods like Isolation Forests. These interpretable alternatives enabled clearer root cause analysis and better alignment with ESG (Environmental, Social, and Governance) reporting requirements. Nevertheless, teams encountered scalability challenges when attempting to embed interpretability into real-time monitoring systems—especially in environments characterized by high-volume or high-dimensional data.
\end{itemize}

Across these domains, the case studies revealed that interpretability is often as much about communication and stakeholder engagement as it is about modeling technique. While no single method universally outperformed others, the combination of case-specific model selection, visualization, and tailored explanation strategies consistently contributed to better stakeholder understanding and trust. These findings reinforce the broader thesis of this whitepaper: \textit{interpretability is a context-dependent practice requiring adaptable frameworks rather than rigid solutions.}

\subsection{Learnings: Practical Adoption of Interpretability}

The collective insights drawn from applied experimentation across sectors demonstrate that interpretability is most effective when treated as a foundational component of the machine learning development process. Embedding interpretability from the outset—starting at the problem definition and model design phase—enables organizations to align techniques with their specific regulatory, operational, and stakeholder needs. Early integration also helps reduce the need for costly and less effective retrofitting later in the development cycle.

Another critical enabler of practical adoption is the development of robust communication infrastructure. Tools such as interactive dashboards, visualizations, and standardized explanation templates help translate complex model behavior into insights that are accessible to diverse stakeholders, including domain experts, auditors, and end users. These tools not only support transparency but also foster trust, a prerequisite for responsible AI deployment.

Interpretability methods must also be carefully selected based on the specific characteristics of the domain and application. There is no one-size-fits-all approach: inherently interpretable models may be better suited for some contexts, while post-hoc techniques might offer flexibility in others. Aligning methods with domain constraints, user literacy, and compliance demands ensures that interpretability delivers meaningful value rather than technical formality.

Finally, organizations benefit from building structured opportunities for reflection, evaluation, and continuous improvement into their workflows. By encouraging teams to document trade-offs, revisit decisions, and gather feedback from end users, interpretability becomes a dynamic capability—capable of adapting as data, models, and requirements evolve.

Together, these recommendations highlight that interpretability is not a fixed property of an algorithm, but a socio-technical practice grounded in context, purpose, and thoughtful design. When supported by the right processes and infrastructure, it becomes a powerful tool for improving accountability, fostering trust, and guiding the responsible adoption of machine learning systems across industries.

\begin{figure}[H]
    \centering
    \includegraphics[scale=0.3]{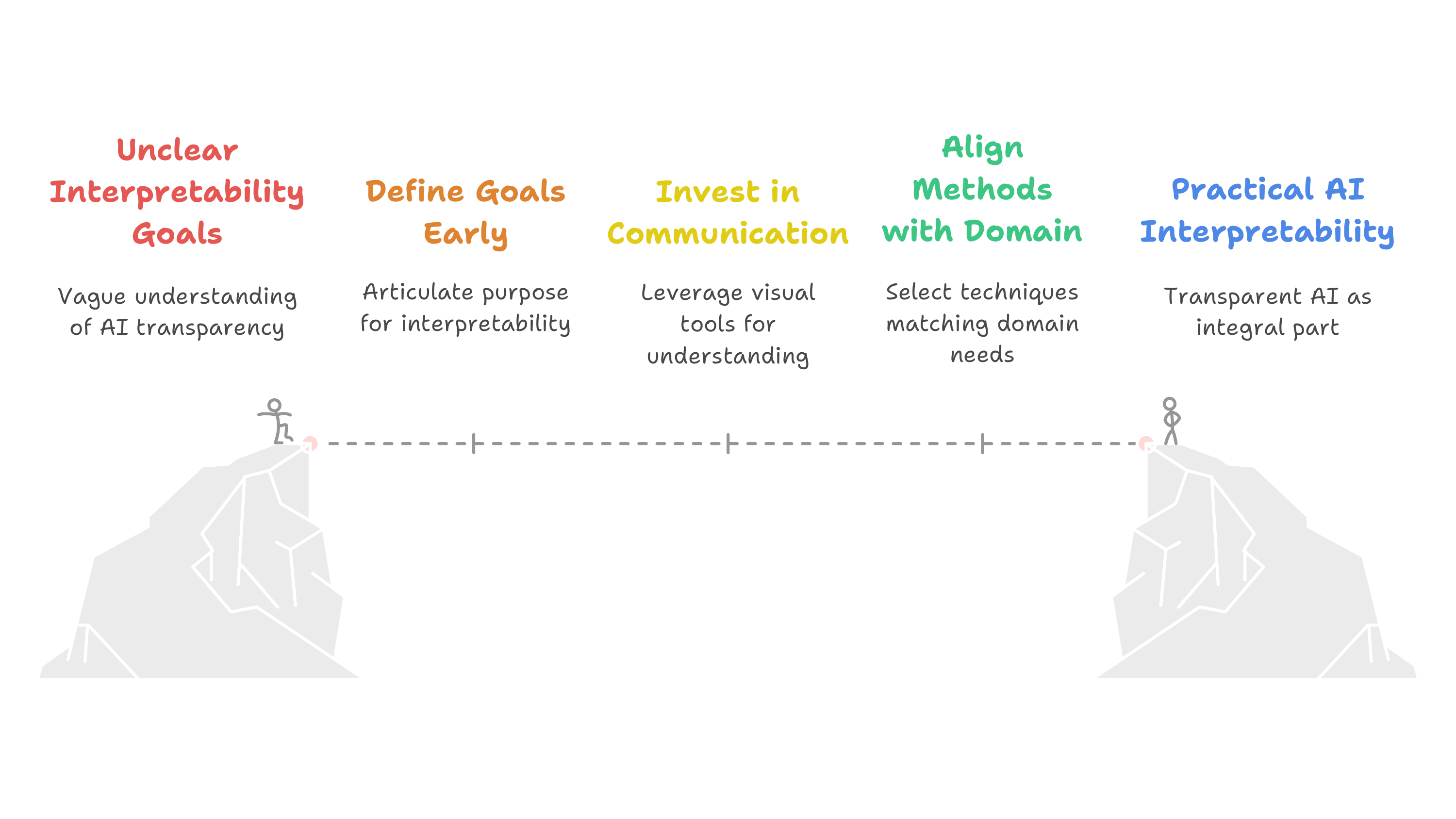}
    \caption{Bridging the Interpretability Gap: Key Steps Toward Practical and Transparent AI Systems}
    \label{fig:learnings}
\end{figure}

\subsection{The Web Portal Initiative: One-Stop Resource for Interpretable AI}

Building on the comprehensive framework presented in this white paper, we propose the development of an integrated web portal that serves as a centralized resource for interpretable AI implementation. This initiative, inspired by the TransparentAI portal concept, would provide practitioners with interactive tools, comprehensive guidance, and real-world case studies to support their interpretability efforts.

The proposed portal would feature several key components designed to address the practical needs identified through our interpretability application experimentation and industry engagement. An interactive technique selection wizard would guide users through a decision tree process to identify the most appropriate interpretability methods based on their specific model types, industry contexts, and stakeholder requirements. This tool would move beyond simple recommendations to provide detailed implementation guidance, including code examples, best practices, and common pitfalls to avoid.

A comprehensive method comparison tool would allow users to evaluate different interpretability approaches side-by-side, with interactive visualizations showing how various techniques perform across different evaluation metrics and use cases. This comparative analysis capability would be particularly valuable for organizations trying to select among multiple viable interpretability approaches for their specific applications.

The portal would include a searchable database of real-world case studies spanning diverse industries and application domains, providing practitioners with concrete examples of how interpretability has been successfully implemented in similar contexts. These case studies would go beyond high-level descriptions to provide detailed technical implementations, lessons learned, and quantified outcomes that demonstrate the business value of interpretable AI.

Interactive evaluation tools would enable users to assess their current interpretability practices against established benchmarks and identify areas for improvement. These tools would provide customized recommendations based on user inputs and guide organizations through a structured improvement process that builds interpretability capabilities over time.

A regulatory compliance checker would help organizations understand and navigate the complex landscape of AI transparency requirements across different jurisdictions and industry sectors. This tool would provide up-to-date information on relevant regulations, assessment frameworks for determining compliance requirements, and guidance on implementing appropriate interpretability measures.

\subsection{Proposed Roadmap for Industry Adoption of Interpretability}
Based on the success and insights gained from interpretability application case studies, we propose a roadmap that organizations can adopt to embed interpretable AI practices into their development workflows and governance structures. This roadmap supports continuous learning, iterative experimentation, and the development of internal capacity to scale interpretable AI across diverse contexts.

\begin{figure}[H]
    \centering
    \includegraphics[scale=0.3]{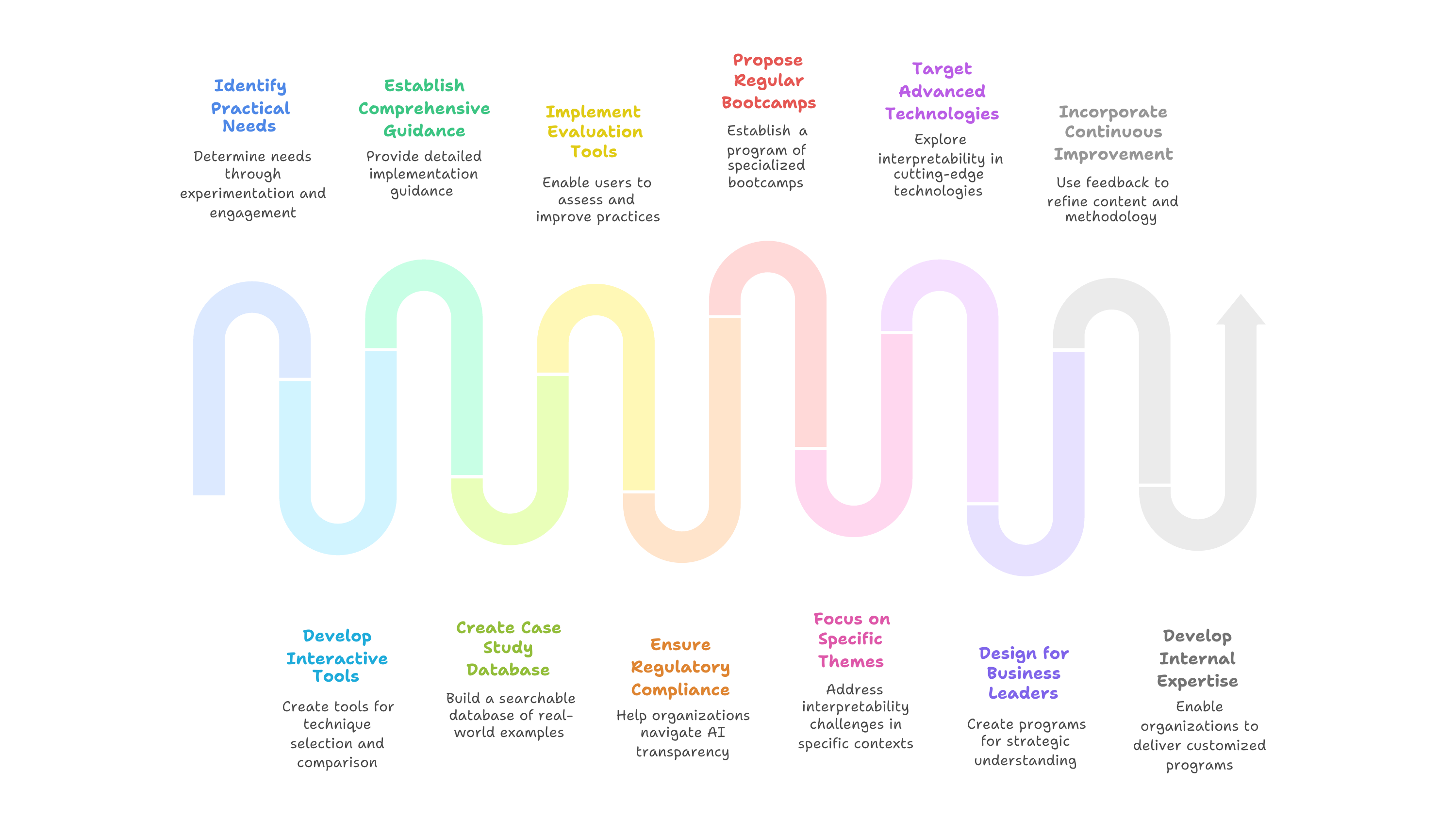}
    \caption{Proposed Roadmap for Driving Industry-Wide Interpretability Adoption}
    \label{fig:roadmap}
\end{figure}

The roadmap encourages organizations to design internal programming such as targeted training, guided pilot projects, and structured knowledge-sharing sessions focused on specific themes or domains (Fig.~\ref{fig:roadmap}). For example, healthcare organizations may prioritize workflows that address unique regulatory, technical, and ethical challenges, while financial institutions might focus on risk management, auditability, and fairness.

Technical workstreams can be established to explore advanced interpretability challenges such as large language models, multi-modal systems, and real-time pipelines. These efforts would focus on evaluating emerging techniques, prototyping interpretable architectures, and identifying tool-chain gaps specific to the organization's use cases.

Leadership focused initiatives should aim to build strategic understanding among business and executive stakeholders. These can include workshops on governance frameworks, change management planning, and the business case for interpretable AI, helping align organizational vision with transparency goals.

Organizations can also foster internal forums or communities of practice to promote cross-functional learning and knowledge transfer. Sharing lessons across departments or use cases can help identify common interpretability challenges and solutions that can be generalized or adapted.

To sustain progress, organizations should incorporate continuous improvement mechanisms into their interpretability initiatives. This includes collecting regular feedback, evaluating progress against defined milestones, and iterating on frameworks, tooling, and training based on real-world implementation outcomes.

Finally, a train-the-trainer strategy can empower internal champions to lead interpretability education across the organization. By developing reusable content, tailored programming, and coaching structures, teams can scale interpretability practices in a sustainable, context-sensitive manner.

This roadmap offers a flexible and actionable foundation for industry adoption of interpretable AI—one that can be adapted to suit varying maturity levels, resource constraints, and sector-specific demands.

\section{Conclusion and Recommendations}

The advancement of AI systems into critical decision-making roles has transformed interpretability from an academic consideration into a fundamental requirement for responsible AI deployment. This whitepaper demonstrates that achieving transparent AI requires coordinated efforts across organizational, regulatory, and human dimensions. Our interpretability application experiment revealed critical insights that reshape how organizations should approach transparent AI implementation. The case studies demonstrated that technical excellence in explanation generation is necessary but insufficient. Teams that achieved meaningful outcomes invested heavily in understanding stakeholder mental models and decision-making workflows. We also propose developing an integrated web portal serving as a centralized resource for interpretable AI implementation, featuring interactive technique selection tools, comprehensive case study databases, and regulatory compliance guidance.

The transition to interpretable AI represents both opportunity and imperative. Organizations should begin by assessing current transparency practices, identifying high-priority applications where interpretability is essential, and developing implementation roadmaps.

The future of AI depends on our collective ability to build systems that humans can understand, trust, and effectively collaborate with. Organizations that embrace interpretability as a fundamental design principle will achieve better regulatory compliance and unlock the full potential of human-AI collaboration. The frameworks presented in this whitepaper provide the foundation—the journey begins with commitment and action.

\printbibliography

@article{hsieh2024,
  title={A Comprehensive Guide to Explainable AI: From Classical Models to LLMs},
  author={Weiche Hsieh and Ziqian Bi and Chuanqi Jiang and Junyu Liu and Benji Peng and Sen Zhang and Xuanhe Pan and Jiawei Xu and Jinlang Wang and Keyu Chen and Pohsun Feng and Yizhu Wen and Xinyuan Song and Tianyang Wang and Ming Liu and Junjie Yang and Ming Li and Bowen Jing and Jintao Ren and Junhao Song and Hong-Ming Tseng and Yichao Zhang and Lawrence K. Q. Yan and Qian Niu and Silin Chen and Yunze Wang and Chia Xin Liang},
  journal={arXiv preprint arXiv:2412.00800},
  year={2024},
  url={https://arxiv.org/abs/2412.00800},
}

@article{luo2024,
  title={From Understanding to Utilization: A Survey on Explainability for Large   Language Models},
  author={Haoyan Luo and Lucia Specia},
  journal={arXiv preprint arXiv:2401.12874},
  year={2024},
  url={https://arxiv.org/abs/2401.12874},
}

@article{kamath2021,
  title={Model Interpretability: Advances in Interpretable Machine Learning},
  author={Uday Kamath and John Liu},
  journal={Explainable Artificial Intelligence: An Introduction to Interpretable Machine Learning},
  year={2021},
  pages={121-165},
  doi={10.1007/978-3-030-83356-5_4},
  url={https://doi.org/10.1007/978-3-030-83356-5_4},
}

@article{černevičienė2024,
  title={Explainable artificial intelligence (XAI) in finance: a systematic literature review},
  author={Jurgita Černevičienė and Audrius Kabašinskas},
  journal={Artificial Intelligence Review},
  year={2024},
  volume={57},
  number={8},
  doi={10.1007/s10462-024-10854-8},
  url={https://doi.org/10.1007/s10462-024-10854-8},
}

@article{han2023ignorance,
  title={Is ignorance bliss? the role of post hoc explanation faithfulness and alignment in model trust in laypeople and domain experts},
  author={Han, Tessa and Ektefaie, Yasha and Farhat, Maha and Zitnik, Marinka and Lakkaraju, Himabindu},
  journal={arXiv preprint arXiv:2312.05690},
  year={2023}
}

@article{jin2022,
  title={Guidelines and Evaluation of Clinical Explainable AI in Medical Image   Analysis},
  author={Weina Jin and Xiaoxiao Li and Mostafa Fatehi and Ghassan Hamarneh},
  journal={arXiv preprint arXiv:2202.10553},
  year={2022},
  url={https://arxiv.org/abs/2202.10553},
}

@article{doshi2017considerations,
  title={Considerations for evaluation and generalization in interpretable machine learning},
  author={Doshi-Velez, Finale and Kim, Been},
  journal={Explainable and interpretable models in computer vision and machine learning},
  pages={3--17},
  year={2017},
  publisher={Springer}
}

@article{nauta2023systematic,
  title={From anecdotal evidence to quantitative evaluation methods: A systematic review on evaluating explainable AI},
  author={Nauta, Meike and Trienes, Jan and Pathak, Shreyasi and Nguyen, Elisa and Peters, Michelle and Schmitt, Yasmin and Schl{\"o}tterer, J{\"o}rg and van Keulen, Maurice and Seifert, Christin},
  journal={ACM Computing Surveys},
  volume={55},
  number={13s},
  pages={1--42},
  year={2023},
  publisher={ACM}
}

@article{velmurugan2024guidelines,
  title={Developing guidelines for functionally-grounded evaluation of explainable artificial intelligence using tabular data},
  author={Velmurugan, Mythreyi and Ouyang, Chun and Xu, Yue and Sindhgatta, Renuka and Wickramanayake, Bemali and Moreira, Catarina},
  journal={Data Mining and Knowledge Discovery},
  year={2024},
  publisher={Springer}
}

@article{Rudin2022,
  title={Interpretable machine learning: Fundamental principles and 10 grand challenges},
  author={Rudin, Cynthia and Chen, Chaofan and Chen, Zhi and Huang, Haiyang and Semenova, Lesia and Zhong, Chudi},
  journal={Statistics Surveys},
  volume={16},
  pages={1--85},
  year={2022},
  publisher={Institute of Mathematical Statistics and Bernoulli Society},
  doi={10.1214/21-SS133},
  url={https://doi.org/10.1214/21-SS133}
}

@article{samek2017evaluating,
  title={Evaluating the visualization of what a deep neural network has learned},
  author={Samek, Wojciech and Binder, Alexander and Montavon, Gr{\'e}goire and Lapuschkin, Sebastian and M{\"u}ller, Klaus-Robert},
  journal={IEEE transactions on neural networks and learning systems},
  volume={28},
  number={11},
  pages={2660--2673},
  year={2017},
  publisher={IEEE}
}

@inproceedings{adebayo2018sanity,
  title={Sanity checks for saliency maps},
  author={Adebayo, Julius and Gilmer, Justin and Muelly, Michael and Goodfellow, Ian and Hardt, Moritz and Kim, Been},
  booktitle={Advances in neural information processing systems},
  volume={31},
  year={2018}
}

@book{molnar2020interpretable,
  title={Interpretable machine learning},
  author={Molnar, Christoph},
  year={2020},
  publisher={Lulu. com}
}

@article{turbe2023evaluation,
  title={Evaluation of post-hoc interpretability methods in time-series classification},
  author={Turb{\'e}, Hugues and Bjelogrlic, Mina and Lovis, Christian and Mengaldo, Gianmarco},
  journal={Nature Machine Intelligence},
  volume={5},
  number={3},
  pages={250--260},
  year={2023},
  publisher={Nature Publishing Group}
}

@article{kenesei2015,
  title={Interpretability of Neural Networks},
  author={Tamás Kenesei and János Abonyi},
  journal={SpringerBriefs in Computer Science},
  year={2015},
  pages={33-48},
  doi={10.1007/978-3-319-21942-4_3},
  url={https://doi.org/10.1007/978-3-319-21942-4_3},
}

\end{document}